\documentclass{article}

\PassOptionsToPackage{numbers, compress}{natbib}

\usepackage[preprint]{neurips_2026}

\usepackage[utf8]{inputenc} 
\usepackage[T1]{fontenc}    
\usepackage{hyperref}       
\usepackage{url}            
\usepackage{booktabs}       
\usepackage{amsfonts}       
\usepackage{amsmath}
\usepackage{nicefrac}       
\usepackage{microtype}      
\usepackage{xcolor}         
\usepackage{graphicx}
\usepackage{algorithm}
\usepackage{algpseudocode}
\usepackage[capitalise]{cleveref} 

\title{Streaming-Aware Diffusion for Real-Time Video Super-Resolution via Cross-Step Attention}

\author{%
  Harris Partaourides \\
  Ethical AI Novelties \\
  Limassol, Cyprus \\
  \texttt{harris.partaourides@ethicalaicy.com} \\
  \And
  Sotirios Chatzis \\
  Cyprus University of Technology \\
  Limassol, Cyprus \\
  \texttt{sotirios.chatzis@cut.ac.cy} \\
}

\begin{document}

\maketitle

\begin{abstract}
Real-time video super-resolution requires high spatio-temporal fidelity under strict latency constraints, challenging diffusion models due to their iterative sampling cost and limited temporal coordination. We propose a streaming-aware framework that adapts pretrained single-image latent diffusion models for efficient video super-resolution (VSR) by exploiting the sequential structure of video streams. Our Cross-Step Attention mechanism reuses intermediate denoising features across adjacent frames and diffusion steps, enabling temporal information exchange without explicit temporal modeling. We further introduce Trajectory-Coupled Diffusion Scheduling, which aligns adjacent diffusion states and provides cleaner intermediate representations for cross-step conditioning, improving temporal coherence. These components are integrated into a streaming inference pipeline that incrementally propagates latent states across frames, reducing the effective computational complexity from $O(N \cdot S)$ to $O(N + S)$ for $N$ frames and $S$ diffusion steps. Experiments on REDS4 and YouHQ40-Test demonstrate improved perceptual quality and temporal realism while maintaining frame-wise stability. Our method achieves over 40 FPS at $512 \times 512$ resolution after cold start, enabling real-time VSR without explicit temporal modeling.

\end{abstract}

\section{Introduction}
\label{sec:intro}

Real-time, high-quality video super-resolution (VSR) is important for streaming applications such as video-on-demand, live broadcasting, video conferencing, and cloud gaming~\cite{caballero2017real, zhang2020improving, zhang2021video}. These applications require both high visual fidelity and low latency, making efficient temporal processing essential~\cite{ignatov2021real, cao2021real}. Diffusion-based single-image super-resolution (SISR) models achieve strong perceptual quality and high-frequency detail synthesis~\cite{saharia2022image, gao2023implicit, podell2023sdxl}, but applying them independently to video frames causes flickering and temporal inconsistency~\cite{xie2023mitigating}. Existing VSR methods address this through motion estimation, optical flow, or recurrent alignment~\cite{wang2024exploiting, rota2024enhancing, zhou2024upscale}, but these mechanisms introduce additional computational and architectural overhead that limits real-time streaming.

We propose a streaming-aware diffusion framework that adapts pretrained SISR diffusion models for efficient and temporally consistent VSR. Our key idea is to exploit the sequential structure shared by video streams and diffusion trajectories. We introduce a model-agnostic \emph{Cross-Step Attention} mechanism that allows the current frame at diffusion step $s$ to attend to intermediate features of the previous frame at step $s+1$. This reuses progressively refined representations across both frames and diffusion steps, providing implicit temporal coordination without explicit motion modeling or additional temporal modules.
To support this interaction, we introduce \emph{Trajectory-Coupled Diffusion Scheduling}, which couples adjacent diffusion states through a vectorized noise schedule. The resulting trajectory-aligned representation exposes cleaner intermediate states from the previous frame as conditioning information during denoising, forming a cross-step clean prior that improves temporal coherence while remaining compatible with standard diffusion training.
We further develop a streaming inference pipeline that incrementally propagates latent states across frames. Instead of independently executing all $S$ diffusion steps for each of $N$ frames, the pipeline maintains a FIFO buffer aligned with the diffusion trajectory and advances the buffered states jointly. This amortizes diffusion computation across the stream and reduces the effective complexity from $O(N \cdot S)$ to $O(N + S)$, enabling real-time inference after the initial pipeline latency.

We evaluate the proposed framework on REDS~\cite{nah2019ntire} and YouHQ40 \cite{zhou2024upscale} using lightweight latent diffusion models~\cite{rombach2022high}. 
Experiments demonstrate improvements in perceptual quality and temporal realism while maintaining frame-wise temporal stability. The streaming implementation achieves over 40 FPS at $512 \times 512$ resolution after cold start. We further validate the model-agnostic design across SDEdit~\cite{meng2021sdedit}, Stable Diffusion~\cite{rombach2022high}, and SDXL~\cite{podell2023sdxl}, showing consistent efficiency gains under streaming inference.

In summary, our contributions are:
\begin{itemize}
\item \textbf{Cross-Step Attention:} A model-agnostic attention mechanism that reuses intermediate diffusion features across adjacent frames and diffusion steps, enabling temporal coordination without explicit temporal modeling.
\item \textbf{Trajectory-Coupled Diffusion Scheduling:} A trajectory-aligned formulation that couples adjacent diffusion states and provides cleaner intermediate representations for cross-step conditioning.
\item \textbf{Streaming-Aware Diffusion Pipeline:} A FIFO-based inference pipeline that amortizes diffusion computation across frames, reducing effective complexity from $O(N \cdot S)$ to $O(N + S)$ and enabling real-time VSR.
\end{itemize}

\section{Related Work}
\label{sec:related_work}

\textbf{Diffusion-based Super-Resolution:}
Diffusion models~\cite{ho2020denoising, dhariwal2021diffusion} have shown strong capability for perceptual super-resolution by synthesizing realistic high-frequency details. SR3~\cite{saharia2022image} performs iterative pixel-space denoising, while latent diffusion models (LDMs) \cite{rombach2022high} reduce the computational cost by operating in latent space. Pretrained models such as Stable Diffusion~\cite{rombach2022high} and SDXL~\cite{podell2023sdxl} have also been adapted for image restoration and super-resolution~\cite{wang2024exploiting}. However, these methods process images independently and do not explicitly address temporal consistency.

\textbf{Diffusion-based Video Super-Resolution:}
VSR requires both spatial detail restoration and temporal consistency. Existing VSR methods typically employ motion estimation, optical flow, recurrent propagation, or feature alignment~\cite{wang2024exploiting, rota2024enhancing, zhou2024upscale}. Recent diffusion-based VSR approaches further introduce temporal attention or multi-frame conditioning, improving consistency at the cost of additional computation. These designs are less suitable for real-time streaming, where frames arrive sequentially under strict latency constraints.

\textbf{Cross-frame Diffusion Feature Reuse.}
Another line of work has demonstrated that pretrained image diffusion models can be adapted to temporally coherent video generation by reusing intermediate diffusion features across frames. Text-to-video diffusion methods~\cite{blattmann2023stable, blattmann2023align, yang2023rerender} and Text2Video-Zero~\cite{khachatryan2023text2video}, for example, exploit cross-frame representations or intermediate diffusion states to establish temporal correspondence without training a fully dedicated video diffusion model. These methods reveal an important property of diffusion models: intermediate denoising representations can serve as effective carriers of temporal information. Our work builds on this principle but introduces a fundamentally different coupling strategy. Instead of reusing features at the same diffusion step, \emph{Cross-Step Attention} connects adjacent video frames at neighboring denoising steps and combines this interaction with \emph{Trajectory-Coupled Diffusion Scheduling}. This design turns the diffusion trajectory itself into a mechanism for causal temporal propagation, while avoiding explicit motion estimation and multi-frame temporal architectures.

\section{Methodology}
\label{sec:methodology}

\begin{figure}[t]
  \centering
  \includegraphics[width=\linewidth]{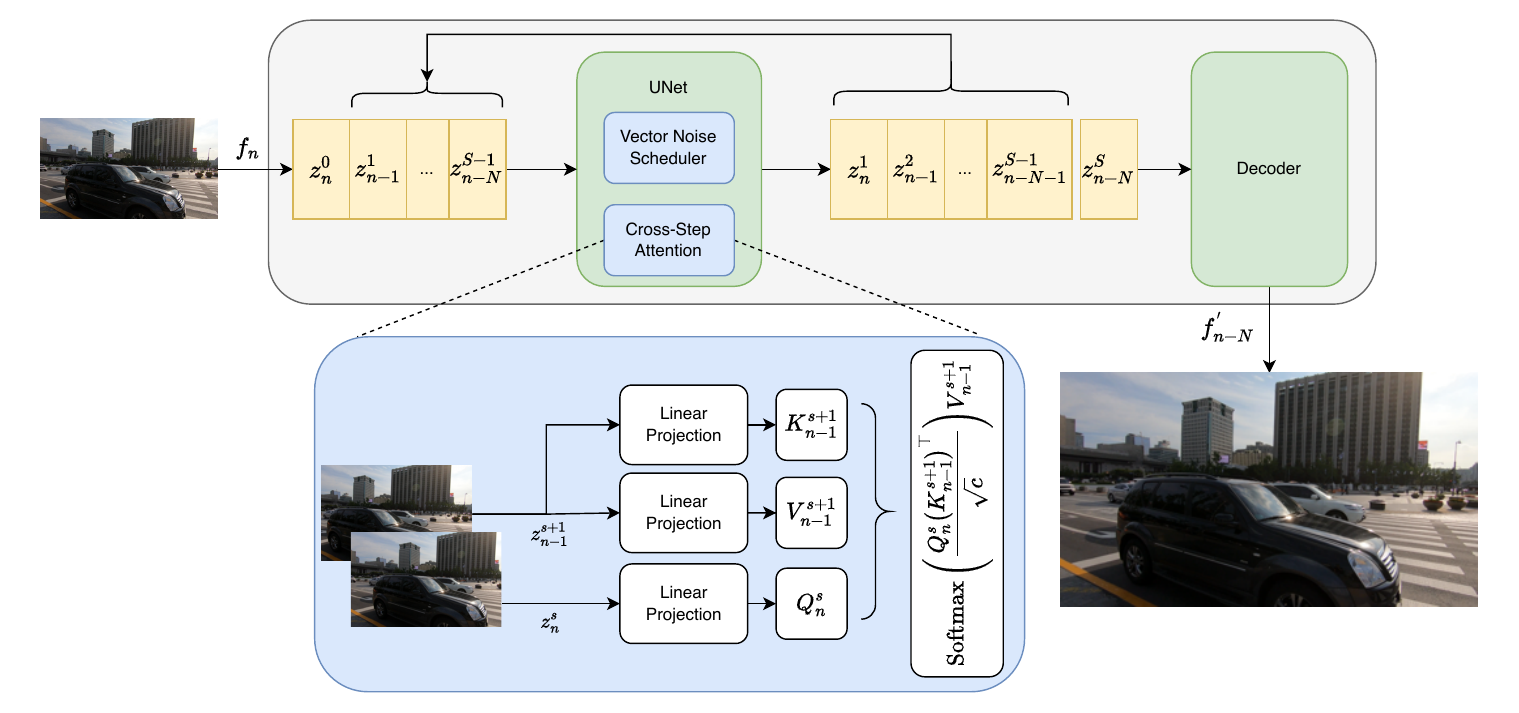}
  \caption{\emph{Cross-Step Attention} for stream-oriented diffusion. Frame $n$ at step $s$ attends to frame $n{-}1$ at step $s{+}1$ for efficient feature reuse and temporal consistency.}
  \label{fig:stream_diffusion}
\end{figure}

Our objective is to develop a stream-oriented diffusion framework for real-time video super-resolution (VSR). As illustrated in Fig.~\ref{fig:stream_diffusion}, we integrate \emph{Cross-Step Attention} into a latent diffusion backbone \cite{ronneberger2015u}, coupling the temporal progression of video frames with the diffusion denoising trajectory. This enables intermediate feature reuse across frames and diffusion steps while avoiding explicit temporal modules.

\subsection{Latent Diffusion Model}

Latent diffusion models (LDMs) improve efficiency by operating in a compressed latent space instead of pixel space. Given an input image $x_0$, a variational autoencoder (VAE) encoder $\mathcal{E}$ maps it to a latent representation:
\begin{equation}
z_0 = \mathcal{E}(x_0).
\end{equation}

A forward diffusion process is then defined in the latent space as:
\begin{equation}
\label{eq:forward}
q(z_s \mid z_{s-1}) = \mathcal{N}(z_s; \sqrt{\alpha_s} z_{s-1}, (1 - \alpha_s) I),
\end{equation}
where $\alpha_s$ is the noise schedule. This yields the closed-form marginal:
\begin{equation}
q(z_s \mid z_0) = \mathcal{N}(z_s; \sqrt{\bar{\alpha}_s} z_0, (1 - \bar{\alpha}_s) I),
\end{equation}
with $\bar{\alpha}_s = \prod_{m=1}^{s} \alpha_m$.

The reverse process is modeled by a neural network $\epsilon_\theta$:
\begin{equation}
p_\theta(z_{s-1} \mid z_s) = \mathcal{N}(z_{s-1}; \mu_\theta(z_s, s), \Sigma),
\end{equation}
where $\mu_\theta$ is parameterized via the predicted noise $\epsilon_\theta(z_s, s)$. The model is trained using the standard denoising objective:
\begin{equation}
L(\theta) = \mathbb{E}_{z_0, \epsilon, s} \left[ \|\epsilon - \epsilon_\theta(z_s, s)\|^2 \right],
\end{equation}
with $\epsilon \sim \mathcal{N}(0, I)$.

The denoised latent is finally mapped back to image space using the VAE decoder $\mathcal{D}$:
\begin{equation}
\hat{x}_0 = \mathcal{D}(z_0).
\end{equation}

Within the UNet backbone of LDMs, self-attention is applied to latent feature maps $Z \in \mathbb{R}^{H \times W \times C}$ by projecting them into queries, keys, and values. Attention is computed as:
\begin{equation}
\mathrm{Attn}(Q, K, V) = \mathrm{Softmax}\left(\frac{QK^\top}{\sqrt{C}}\right)V.
\end{equation}

This mechanism enables long-range spatial interactions and improves detail synthesis. However, in standard LDMs, self-attention is applied independently to each image and diffusion timestep, without any temporal interaction across frames. As a result, they are not inherently designed for video generation or temporal consistency.

\subsection{Cross-Step Streaming Attention}

To address this limitation, we introduce a \emph{Cross-Step Attention} mechanism that enables structured information exchange across frames and diffusion timesteps in a streaming setting. By integrating this module into the self-attention layers of a pretrained SISR diffusion model, we obtain a causal and plug-and-play adaptation for video while preserving the sequential nature of both diffusion denoising and streaming inference.

Our design is inspired by cross-frame attention mechanisms in Text2Video-Zero~\cite{khachatryan2023text2video}, but differs in two key aspects. First, we operate in a strictly streaming setting, where each frame only attends to the immediately preceding frame rather than a global reference frame. At frame index $n$, standard self-attention is replaced by cross-frame attention between the current latent $z_n$ and the previous latent $z_{n-1}$:
\begin{equation}
Z_n = \left[ z_n, z_{n-1} \right] \in \mathbb{R}^{2 \times H \times W \times C},
\end{equation}
where queries are computed from $z_n$, while keys and values are computed from $z_{n-1}$. This yields:
\begin{equation}
\mathrm{CF\text{-}Attn}(Q_n, K_{n-1}, V_{n-1}) =
\mathrm{Softmax}\left(\frac{Q_n K_{n-1}^\top}{\sqrt{C}}\right)V_{n-1},
\end{equation}
with standard self-attention recovered for $n=1$.

Second, we extend this mechanism along the diffusion trajectory by introducing \emph{Cross-Step Attention}. Instead of attending within the same diffusion timestep, we exploit the sequential nature of the denoising trajectory and let the current frame at step $s$ attend to the previous frame further along the denoising trajectory at step $s+1$:
\begin{equation}
Z_n^{s} = \left[ z_n^{s}, z_{n-1}^{s+1} \right] \in \mathbb{R}^{2 \times H \times W \times C}.
\end{equation}
This produces queries from $z_n^{s}$ and keys/values from $z_{n-1}^{s+1}$, leading to:
\begin{equation}
\label{eq:cross-step-attn}
\mathrm{CS\text{-}Attn}(Q_n^s, K_{n-1}^{s+1}, V_{n-1}^{s+1}) =
\mathrm{Softmax}\left(\frac{Q_n^s (K_{n-1}^{s+1})^\top}{\sqrt{C}}\right)V_{n-1}^{s+1}.
\end{equation}

By leveraging intermediate denoising states further along the denoising trajectory of the previous frame, the model reuses progressively refined structural information to guide the current denoising process. This enforces temporal consistency while avoiding redundant computation across both frames and diffusion steps, leading to improved stability and efficiency in streaming video generation.
This formulation naturally interacts with the denoising trajectory, motivating a \emph{Trajectory-Coupled Diffusion Scheduling} strategy that aligns intermediate states across diffusion steps for both training and inference.

\subsection{Trajectory-Coupled Diffusion Scheduling}
\label{sec:trajectory_coupled_diffusion}

To support \emph{Cross-Step Attention} in a streaming setting, we introduce a \emph{Trajectory-Coupled Diffusion Scheduling} formulation that models local dependencies along the denoising trajectory. Instead of treating each timestep independently, we lift the scalar noise schedule into a paired representation over adjacent diffusion steps.

Given a standard scalar noise schedule $\{\alpha_s\}_{s=1}^{S}$, we construct a vectorized schedule by stacking consecutive values:
\begin{equation}
\boldsymbol{\alpha}_s =
\begin{bmatrix}
\alpha_s \\
\alpha_{s+1}
\end{bmatrix},
\end{equation}
which explicitly couples adjacent diffusion steps $s$ and $s+1$.

This induces a corresponding paired latent representation along the denoising trajectory:
\begin{equation}
\tilde{\mathbf{z}}_n^{s} = (z_n^{s}, z_n^{s+1}),
\end{equation}
enabling simultaneous access to two consecutive states within the same trajectory without modifying the underlying diffusion process.

We construct trajectory-aligned training pairs:
\begin{equation}
\mathcal{P} = \{(z_n^{s}, z_{n-1}^{s+1})\},
\end{equation}
aligning the current frame at step $s$ with the previous frame at a more advanced step $s+1$. This induces a cross-step clean prior, where $z_{n-1}^{s+1}$ provides a refined denoising signal for $z_n^{s}$. During training, the conditioning state $z_{n-1}^{s+1}$ is obtained by applying the forward process of \cref{eq:forward} to the previous frame at level $s+1$, whereas at inference it is the model-produced latent already present in the streaming buffer at that diffusion depth. Because keys and values originate from a more advanced (cleaner) step than the query, the conditioning distribution remains close across the two regimes, which limits the train--inference mismatch in practice.

The model is trained using the standard diffusion denoising objective:
\begin{equation}
\mathcal{L}(\theta) =
\mathbb{E}_{(z_n^{s}, z_{n-1}^{s+1}), \epsilon}
\left[
\left\| \epsilon - \epsilon_\theta(z_n^{s}, z_{n-1}^{s+1}) \right\|^2
\right],
\end{equation}
with $\epsilon \sim \mathcal{N}(0, I)$.

At inference, this formulation enables consistent access to coupled states $\tilde{\mathbf{z}}_n^{s} = (z_n^{s}, z_n^{s+1})$, reducing redundancy across diffusion steps and frames while complementing \emph{Cross-Step Attention}. This structured coupling naturally leads to a unified streaming inference formulation, described next.

\subsection{Streaming-Aware Diffusion for Video Super-Resolution}

Our streaming inference pipeline builds on \emph{Cross-Step Attention} and the trajectory-coupled diffusion formulation, enabling structured reuse of intermediate denoising states across both diffusion timesteps and video frames.

Real-time video super-resolution requires aligning computational throughput with the incoming frame rate. However, even efficient diffusion models struggle under high FPS or resource-constrained settings, making it necessary to decouple frame arrival from full diffusion completion through amortized computation.
In standard SISR-based diffusion, each frame is independently denoised over $S$ steps, leading to a computational cost of $O(N \cdot S)$ for a sequence of $N$ frames due to the lack of temporal state reuse.

To address this, we reformulate inference as a streaming FIFO pipeline with buffer size $S$, matching the diffusion horizon. At time $n$, the system maintains a structured latent state:
\begin{equation}
Z_n = \{ z_{n-k}^{k} \mid k = 1,\dots,S \},
\end{equation}
where each element corresponds to a frame progressing through a specific diffusion depth.
Each incoming frame is encoded into its initial latent state and enters the buffer at diffusion step $0$. At every iteration, all latents advance by one diffusion step under the trajectory-coupled denoising process with \emph{Cross-Step Attention}. Consequently, each frame traverses the buffer over $S$ updates before being decoded at the final stage.

This induces a pipeline latency of $S$ steps, after which the system produces one output frame per iteration in steady state. The resulting process transforms diffusion from independent per-frame generation into a continuous state evolution over a shared denoising trajectory.
Computationally, each diffusion update is executed once per time step rather than $S$ times per frame, yielding amortized complexity: $ O(N + S)$.
The FIFO-aligned design ensures temporal consistency, efficient state reuse, and full compatibility with the underlying diffusion process.


\section{Experiments}
\label{sec:experiments}

We evaluate the proposed streaming-aware diffusion framework on two complementary VSR benchmarks: REDS4~\cite{nah2019ntire} and YouHQ40-Test~\cite{zhou2024upscale} enabling evaluation of reconstruction quality, perceptual quality, temporal consistency, and streaming efficiency.

\textbf{REDS Dataset:}
The REDS dataset~\cite{nah2019ntire} is a standard benchmark for video restoration tasks, consisting of high-quality video sequences with diverse motion patterns. Following prior work, we evaluate on the REDS4 subset, which contains four commonly used test sequences (000, 011, 015, and 020). We adopt the standard $4\times$ super-resolution setting using the provided low-resolution sequences ($320\times180 \rightarrow 1280\times720$). For fine-tuning, we use the official REDS validation split, which contains 30 video sequences.

\textbf{YouHQ40-Test Set:}
YouHQ40-Test~\cite{zhou2024upscale} is a real-world high-quality VSR benchmark derived from the YouHQ dataset. It consists of 40 video clips (approximately 32 frames each), covering diverse scenes including urban environments, natural landscapes, human activities, underwater scenes, and low-light conditions. The dataset includes both $1920\times1080$ (Full HD) and $1080\times1080$ (square) videos. For evaluation, we apply $4\times$ super-resolution on $1080\times1080$ videos ($270\times270 \rightarrow 1080\times1080$). For $1920\times1080$ videos, we extract a center-cropped $1080\times1080$ region and apply the same degradation setting for consistency.

\textbf{Implementation details:}
We implement our method in PyTorch using the Diffusers library. \emph{Cross-Step Attention} is implemented as a modular attention processor compatible with pretrained latent diffusion backbones, supporting both \texttt{AttnProcessor} and \texttt{AttnProcessor2\_0}. Experiments are conducted on a single NVIDIA RTX 4090 GPU under identical inference settings. We fine-tune only the attention-related modules while keeping the pretrained diffusion backbone fixed. Streaming throughput is measured after the initial cold-start period.

\textbf{Evaluation Metrics:}
We evaluate performance using spatial fidelity, perceptual quality, and temporal consistency metrics. Following the perception--distortion trade-off~\cite{blau2018perception}, we emphasize perceptual metrics while reporting distortion-based measures for completeness.
For spatial fidelity, we report Peak Signal-to-Noise Ratio (PSNR) and Structural Similarity Index (SSIM). To better capture perceptual quality, we use Learned Perceptual Image Patch Similarity (LPIPS)~\cite{zhang2018unreasonable} and Deep Image Structure and Texture Similarity (DISTS)~\cite{ding2020image}. We also report Fr\'echet Inception Distance (FID)~\cite{salimans2016improved} as a distribution-level metric computed over extracted frame features.
For temporal consistency, we report Fr\'echet Video Distance (FVD)~\cite{unterthiner2018towards}, which measures distributional similarity in a spatio-temporal feature space. In addition, we use temporal LPIPS (tLPIPS) and temporal DISTS (tDISTS) to measure frame-to-frame perceptual stability. Following prior work~\cite{rota2024enhancing, adiya2024bidirectional}, temporal LPIPS is defined as the absolute change in perceptual distance between consecutive frames:
\begin{equation}
\text{tLPIPS}_t = \left| \text{LPIPS}(f_t, f_{t+1}) - \text{LPIPS}(f_t^\ast, f_{t+1}^\ast) \right|,
\end{equation}
where $f_t$ and $f_t^\ast$ denote ground-truth and generated frames, respectively. tDISTS is defined analogously by replacing LPIPS with DISTS.

\subsection{Cross-Step Attention Evaluation}

We evaluate \emph{Cross-Step Attention} using an LDM, denoted as $\text{LDM}_{\text{CS}}$, on REDS4 and YouHQ40-Test with 30 diffusion steps. We compare against bicubic interpolation, vanilla LDM, Stable-VSR, and Real-ESRGAN. Table~\ref{tab:merged} reports spatial reconstruction, perceptual, and temporal metrics. For the fine-tuned variant $\text{LDM}_{\text{CS}}^{*}$, we fine-tune the attention-related parameters on the REDS validation set for 15 epochs using Adam with a learning rate of $1\times10^{-5}$. The resulting model is evaluated directly on both REDS4 and YouHQ40-Test without additional adaptation.

\begin{table}[t]
\small
\centering
\caption{Quantitative comparison on REDS4 and YouHQ40-Test with 30 diffusion steps. $\text{LDM}_{\text{CS}}$ denotes \emph{Cross-Step Attention}, while $\text{LDM}_{\text{CS}}^{*}$ is fine-tuned on REDS for 15 epochs.}
\label{tab:merged}
\setlength{\tabcolsep}{4pt}
\begin{tabular}{lcccccccc}
\toprule
\textbf{Method} & \textbf{PSNR} & \textbf{SSIM} & \textbf{LPIPS} & \textbf{DISTS} & \textbf{FID} &
\textbf{tLPIPS} &
\textbf{tDISTS} &
\textbf{FVD} \\
\midrule

\multicolumn{9}{c}{\textbf{REDS4}} \\
\midrule
Bicubic & 26.13 & 0.73 & 0.453 & 0.186 & 5.80 & 0.023 & 0.003 & 18.89 \\
Real-ESRGAN & 23.35 & 0.67 & 0.242 & 0.105 & 2.97 & 0.006 & 0.007 & 23.48 \\
Stable-VSR & 28.15 & 0.80 & 0.101 & 0.046 & 0.06 & 0.004 & 0.006 & 2.59 \\
LDM & 24.66 & 0.69 & 0.214 & 0.097 & 2.20 & 0.030 & 0.025 & 11.71 \\
$\text{LDM}_{\text{CS}}$ & 24.07 & 0.65 & 0.198 & 0.092 & 1.43 & 0.044 & 0.027 & 8.84 \\
$\text{LDM}_{\text{CS}}^{*}$ & 25.26 & 0.69 & 0.180 & 0.083 & 0.84 & 0.028 & 0.023 & 7.30 \\
\midrule

\multicolumn{9}{c}{\textbf{YouHQ40-Test}} \\
\midrule

Bicubic        & 27.60 & 0.78 & 0.365 & 0.163 & 2.20 & 0.019 & 0.004 & 73.41 \\
Real-ESRGAN    & 23.80 & 0.68 & 0.269 & 0.137 & 2.79 & 0.007 & 0.007 & 71.02 \\
Stable-VSR     & 27.58 & 0.78 & 0.175 & 0.087 & 0.17 & 0.009 & 0.006 & 64.19 \\
LDM            & 25.25 & 0.70 & 0.241 & 0.123 & 1.53 & 0.059 & 0.031 & 66.73 \\
$\text{LDM}_{\text{CS}}$ & 24.42 & 0.65 & 0.240 & 0.126 & 1.30 & 0.093 & 0.038 & 65.27 \\
$\text{LDM}_{\text{CS}}^{*}$ & 26.12 & 0.72 & 0.199 & 0.107 & 0.35 & 0.060 & 0.029 & 65.63 \\
\bottomrule
\end{tabular}
\end{table}

The results reveal a clear distinction between the zero-shot and fine-tuned settings. In the zero-shot setting, $\text{LDM}_{\text{CS}}$ improves perceptual and distribution-level temporal metrics over vanilla LDM, although this comes with a degradation in distortion-based and frame-wise temporal metrics. On REDS4, LPIPS decreases from 0.214 to 0.198 and FID from 2.20 to 1.43, while FVD decreases from 11.71 to 8.84. Similarly, on YouHQ40-Test, FID decreases from 1.53 to 1.30 and FVD from 66.73 to 65.27. These improvements indicate that cross-step feature reuse can improve perceptual and distribution-level temporal quality without task-specific training. However, PSNR/SSIM decrease and frame-wise temporal errors increase, particularly on YouHQ40-Test. This behavior is consistent with the perception--distortion trade-off~\cite{blau2018perception}, where improved perceptual realism can occur at the expense of pixel-level fidelity and local frame-to-frame stability.

Fine-tuning substantially improves this trade-off. On REDS4, $\text{LDM}_{\text{CS}}^{*}$ improves PSNR from 24.66 to 25.26 relative to vanilla LDM, while reducing LPIPS from 0.214 to 0.180, DISTS from 0.097 to 0.083, and FID from 2.20 to 0.84. FVD also decreases from 11.71 to 7.30. On YouHQ40-Test, PSNR increases from 25.25 to 26.12, while LPIPS, DISTS, and FID decrease from 0.241, 0.123, and 1.53 to 0.199, 0.107, and 0.35, respectively. The improvement transfers to YouHQ40-Test despite training only on REDS, suggesting that the learned cross-step representations generalize beyond the fine-tuning domain. Importantly, fine-tuning also mitigates the frame-wise temporal instability introduced by the zero-shot model: on REDS4, tLPIPS/tDISTS improve from 0.044/0.027 for $\text{LDM}_{\text{CS}}$ to 0.028/0.023 for $\text{LDM}_{\text{CS}}^{*}$, approaching or improving upon the corresponding vanilla LDM values. On YouHQ40-Test, the temporal metrics remain essentially unchanged relative to vanilla LDM while the perceptual metrics improve substantially. Thus, lightweight fine-tuning provides a more favorable operating point that retains the perceptual benefits of \emph{Cross-Step Attention} while largely recovering frame-wise temporal stability.

The comparison with existing methods further highlights the quality-efficiency trade-off. Stable-VSR achieves the strongest reconstruction and perceptual quality among the compared methods, benefiting from explicit temporal modeling. However, this comes with substantially higher computational cost (see \cref{sec:efficiency}). Real-ESRGAN provides efficient inference but achieves weaker perceptual quality, while bicubic interpolation serves as a simple reconstruction baseline. In contrast, $\text{LDM}_{\text{CS}}^{*}$ provides a lightweight alternative that improves the perceptual quality of a pretrained diffusion backbone without introducing a dedicated temporal architecture.

\begin{figure}[t]
  \centering
  \begin{minipage}{\linewidth}
    \centering
    \includegraphics[width=\linewidth]{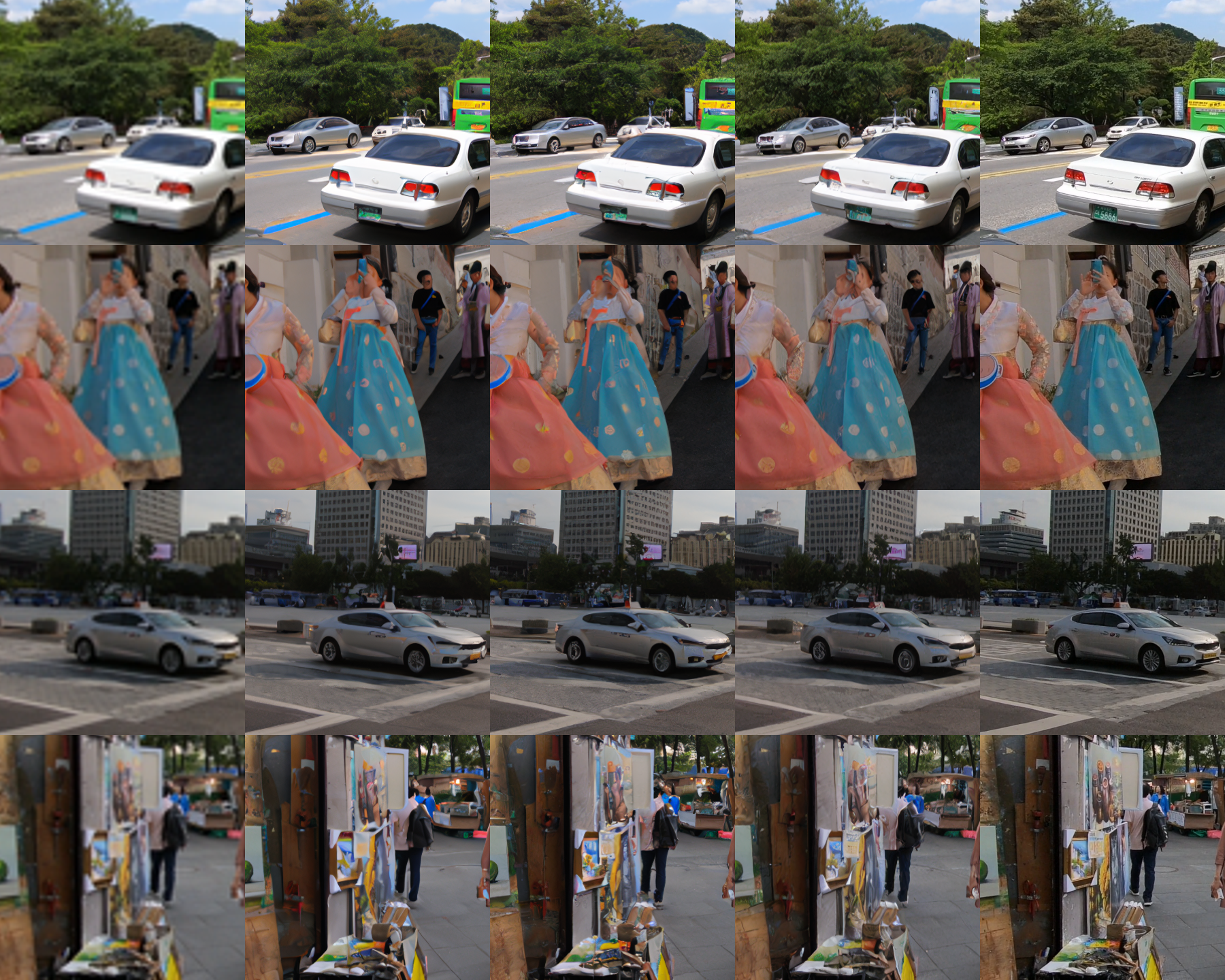}

    \vspace{1mm}
    \makebox[\linewidth][c]{%
      \makebox[0.2\linewidth]{(a) Bicubic}%
      \makebox[0.2\linewidth]{(b) LDM}%
      \makebox[0.2\linewidth]{(c) $LDM_{CS}$}%
      \makebox[0.2\linewidth]{(d) $LDM_{CS}^*$}%
      \makebox[0.2\linewidth]{(e) High-Res}%
    }
  \end{minipage}
  \vspace{1pt}
  \caption{REDS4 examples with different methods.}
  \label{fig:Cross_step}
\end{figure}

Qualitative comparisons are shown in Fig.~\ref{fig:Cross_step}. $\text{LDM}_{\text{CS}}^{*}$ produces sharper details and more coherent structures than vanilla LDM, particularly around object boundaries, repetitive patterns, and fine-grained textures. In the first row, repetitive structures on the car and bus are reconstructed with improved geometric regularity and fewer visible distortions. In the second row, the dress exhibits more consistent appearance and fewer local artifacts. In the third row, fine details on the moving vehicle are better preserved, with less visible blurring in high-frequency regions. In the final row, object boundaries between the window and pedestrian regions appear cleaner, with reduced foreground--background ambiguity. These visual differences are consistent with the improvements in LPIPS, DISTS, and FID reported in Table~\ref{tab:merged}.

Overall, the results show that \emph{Cross-Step Attention} provides a useful mechanism for transferring intermediate diffusion representations across adjacent frames. In the zero-shot setting, this primarily improves perceptual and distribution-level temporal quality, while lightweight fine-tuning substantially reduces the associated loss in frame-wise temporal stability. The resulting approach therefore provides a practical way to introduce temporal feature reuse into pretrained diffusion restoration models without explicit motion estimation or a dedicated multi-frame temporal architecture.

\subsection{Efficiency Analysis for Real-Time Video Super-Resolution}
\label{sec:efficiency}

We evaluate inference efficiency at $512^2$ and $1024^2$ resolutions using LDM, SDx4, SDEdit, and SDXL, with and without \emph{Cross-Step Attention} and streaming inference. We additionally compare against bicubic interpolation, Real-ESRGAN, and StableVSR. All diffusion-based methods use 30 sampling steps. We report runtime per 1-second video, diffusion throughput, and output FPS in Table~\ref{tab:efficiency}.

Standard diffusion models remain computationally expensive for video processing. At $512^2$, LDM, SDx4, SDEdit, and SDXL achieve only 2.2, 1.0, 1.4, and 0.8 FPS, respectively. Adding CS attention alone does not improve throughput, since diffusion is still performed independently for each frame. In contrast, the streaming pipeline reorganizes the same denoising updates into a continuous execution schedule, allowing intermediate states to be reused across consecutive frames.

The resulting streaming variants substantially increase throughput. At $512^2$, $\text{LDM}_{\text{CS}}^{\text{stream}}$, $\text{SDx4}_{\text{CS}}^{\text{stream}}$, $\text{SDEdit}_{\text{CS}}^{\text{stream}}$, and $\text{SDXL}_{\text{CS}}^{\text{stream}}$ achieve 44, 17, 34, and 17 FPS, respectively. At $1024^2$, they achieve 12, 4.5, 9, and 7.5 FPS. These results show that the main efficiency gain comes from streaming reorganization, which amortizes diffusion computation across consecutive frames.

StableVSR provides strong reconstruction quality but requires 120~s per 1-second video at $512^2$, compared with only 0.7~s for $\text{LDM}_{\text{CS}}^{\text{stream}}$. Bicubic and Real-ESRGAN remain faster in some settings, but do not provide comparable diffusion-based perceptual reconstruction. Together with the quality results in Table~\ref{tab:merged}, these measurements demonstrate the practical efficiency advantage of the proposed streaming formulation.

The streaming and non-streaming variants use identical model weights and denoising updates, with the streaming buffer size matched to the number of diffusion steps. To verify that streaming does not alter reconstruction behavior, we compare $\text{LDM}_{\text{CS}}^{\text{stream}}$ with $\text{LDM}_{\text{CS}}$. At $512^2$, the streaming and non-streaming variants obtain FID/FVD values of 0.84/7.47 and 0.84/7.30 on REDS4, and 0.35/65.67 and 0.35/65.63 on YouHQ40-Test, respectively. Peak diffusion process memory is also nearly unchanged, measuring 4.98/4.96~GB at $512^2$ and 18.54/18.50~GB at $1024^2$ for streaming/non-streaming inference. These results indicate that streaming primarily changes the execution order while preserving the underlying denoising computation.

To further assess potential error propagation during streaming, we compare performance across the first, middle, and final thirds of the REDS4 sequences. Temporal quality does not degrade over time; instead, FVD decreases from 9.95 to 7.95 to 7.52, while tLPIPS decreases from 0.030 to 0.028 to 0.026 and tDISTS from 0.024 to 0.023 to 0.023, with spatial quality remaining largely stable. These results suggest that the causal one-hop cross-step design does not introduce accumulating temporal errors during streaming.

\begin{table}[t]
\centering
\small
\renewcommand{\arraystretch}{1.0}
\caption{Inference efficiency at $512^2$ and $1024^2$ resolutions.}
\label{tab:efficiency}
\begin{tabular}{l|cc|cc|cc}
\toprule
\textbf{Method} & \multicolumn{2}{c|}{\textbf{Runtime} $\downarrow$}
& \multicolumn{2}{c|}{\textbf{Steps/s} $\uparrow$}
& \multicolumn{2}{c}{\textbf{FPS} $\uparrow$} \\

& 512$^2$ & 1024$^2$
& 512$^2$ & 1024$^2$
& 512$^2$ & 1024$^2$ \\
\midrule
Bicubic & 1.5 & 3 & -- & -- & 20 & 10 \\
Real-ESRGAN & 3.3 & 9.1 & -- & -- & 9 & 3.3 \\
StableVSR & 120 & 750 & -- & -- & -- & -- \\
\midrule
LDM & 13 & 36 & 67 & 25 & 2.2 & 0.8 \\
$\text{LDM}_{\text{CS}}$ & 20 & 75 & 44 & 12 & 1.5 & 0.4 \\
$\text{LDM}_{\text{CS}}^{\text{stream}}$ & 0.7 & 2.5 & 44 & 12 & 44 & 12 \\
\midrule
SDx4 & 30 & 100 & 30 & 9 & 1.0 & 0.3 \\
$\text{SDx4}_{\text{CS}}$ & 53 & 200 & 17 & 4.5 & 0.6 & 0.2 \\
$\text{SDx4}_{\text{CS}}^{\text{stream}}$ & 2 & 7 & 17 & 4.5 & 17 & 4.5 \\
\midrule
SDEdit & 21 & 56 & 42 & 16 & 1.4 & 0.5 \\
$\text{SDEdit}_{\text{CS}}$ & 26 & 100 & 34 & 9 & 1.1 & 0.3 \\
$\text{SDEdit}_{\text{CS}}^{\text{stream}}$ & 1 & 3 & 34 & 9 & 34 & 9 \\
\midrule
SDXL & 38 & 64 & 24 & 14 & 0.8 & 0.5 \\
$\text{SDXL}_{\text{CS}}$ & 53 & 120 & 17 & 7.5 & 0.6 & 0.3 \\
$\text{SDXL}_{\text{CS}}^{\text{stream}}$ & 2 & 4 & 17 & 7.5 & 17 & 7.5 \\
\bottomrule
\end{tabular}
\end{table}

\subsection{Effect of Diffusion Steps on Reconstruction Performance}

Table~\ref{tab:merged_ablation} examines how the diffusion-step budget affects perceptual and temporal quality on REDS4 and YouHQ40-Test. Increasing the number of diffusion steps consistently improves the vanilla LDM, with the largest gains occurring at low step budgets and diminishing improvements at higher budgets. \emph{Cross-Step Attention} achieves comparable or better perceptual quality at substantially fewer steps, indicating improved sample efficiency of the diffusion process.

On REDS4, the zero-shot $\text{LDM}_{CS}$ already provides substantial gains at low and moderate step budgets. At 10 steps, $\text{LDM}_{CS_{10}}$ achieves an LPIPS of 0.213, comparable to the 0.214 obtained by the 30-step vanilla LDM, while using only one-third of the denoising steps. Its DISTS and FID, however, remain slightly higher at 0.100 and 2.75 compared with 0.097 and 2.20 for $LDM_{30}$. The benefit becomes more pronounced after fine-tuning. $\text{LDM}_{CS_{10}}^{*}$ achieves LPIPS/DISTS/FID/FVD of 0.199/0.091/2.00/10.15, improving over the 30-step vanilla LDM at 0.214/0.097/2.20/11.71 despite using only 10 diffusion steps.
On YouHQ40-Test, the zero-shot gains are more modest, particularly at low step budgets. Fine-tuning substantially improves the perceptual metrics: $\text{LDM}_{CS_{10}}^{*}$ achieves LPIPS, DISTS, and FID of 0.212, 0.113, and 0.94, compared with 0.241, 0.123, and 1.53 for $LDM_{30}$. FVD is 68.76 compared with 66.73 for the 30-step baseline, indicating that the reduction in diffusion steps comes with a small degradation in distribution-level temporal quality on this more diverse dataset. Increasing the step budget progressively reduces this gap, with $\text{LDM}_{CS_{30}}^{*}$ achieving the best results at 0.199 LPIPS, 0.107 DISTS, 0.35 FID, and 65.63 FVD.

These results demonstrate that \emph{Cross-Step Attention} can substantially improve the sample efficiency of diffusion-based VSR. In particular, the fine-tuned 10-step model on REDS4 matches or exceeds the perceptual and temporal quality of the 30-step vanilla LDM while requiring only one-third of the denoising steps. On YouHQ40-Test, the same reduction in steps preserves strong perceptual gains but introduces a modest temporal-quality trade-off, which is progressively reduced as more diffusion steps are used.

\begin{table}[t]
\small
\centering
\caption{Effect of diffusion step budgets on reconstruction and temporal quality.}
\label{tab:merged_ablation}
\setlength{\tabcolsep}{4pt}
\begin{tabular}{l|cccc|cccc}
\toprule
\textbf{Method} & \multicolumn{4}{c|}{\textbf{REDS4}} & \multicolumn{4}{c}{\textbf{YouHQ40-Test}} \\

& LPIPS$\downarrow$ & DISTS$\downarrow$ & FID$\downarrow$ & FVD$\downarrow$
& LPIPS$\downarrow$ & DISTS$\downarrow$ & FID$\downarrow$ & FVD$\downarrow$ \\
\midrule

$LDM_{5}$  
& 0.300 & 0.144 & 5.72 & 21.84 
& 0.302 & 0.154 & 3.96 & 76.10 \\

$LDM_{10}$ 
& 0.263 & 0.121 & 3.74 & 16.44 
& 0.275 & 0.137 & 2.31 & 71.06 \\

$LDM_{15}$ 
& 0.240 & 0.109 & 2.98 & 14.20 
& 0.259 & 0.130 & 1.91 & 69.17 \\

$LDM_{20}$ 
& 0.228 & 0.103 & 2.61 & 13.18 
& 0.250 & 0.126 & 1.71 & 67.95 \\

$LDM_{25}$ 
& 0.220 & 0.100 & 2.35 & 12.21 
& 0.245 & 0.124 & 1.61 & 67.30 \\

$LDM_{30}$ 
& 0.214 & 0.097 & 2.20 & 11.71 
& 0.241 & 0.123 & 1.53 & 66.73 \\

\midrule

$LDM_{CS_{5}}$  
& 0.253 & 0.126 & 5.85 & 19.90 
& 0.289 & 0.155 & 4.96 & 76.96 \\

$LDM_{CS_{10}}$ 
& 0.213 & 0.100 & 2.75 & 12.77 
& 0.244 & 0.130 & 2.00 & 68.70 \\

$LDM_{CS_{15}}$ 
& 0.202 & 0.095 & 1.99 & 11.62 
& 0.239 & 0.127 & 1.52 & 66.69 \\

$LDM_{CS_{20}}$ 
& 0.199 & 0.093 & 1.69 & 9.93 
& 0.237 & 0.126 & 1.38 & 65.86 \\

$LDM_{CS_{25}}$ 
& 0.198 & 0.092 & 1.52 & 9.66 
& 0.238 & 0.125 & 1.32 & 65.10 \\

$LDM_{CS_{30}}$ 
& 0.197 & 0.092 & 1.30 & 8.84 
& 0.240 & 0.126 & 1.30 & 65.27 \\

\midrule

$LDM_{CS_{5}}^{*}$  
& 0.232 & 0.107 & 4.21 & 15.26 
& 0.236 & 0.124 & 2.45 & 73.36 \\

$LDM_{CS_{10}}^{*}$ 
& 0.199 & 0.091 & 2.00 & 10.15 
& 0.212 & 0.113 & 0.94 & 68.76 \\

$LDM_{CS_{15}}^{*}$ 
& 0.188 & 0.086 & 1.38 & 8.69 
& 0.203 & 0.109 & 0.60 & 67.37 \\

$LDM_{CS_{20}}^{*}$ 
& 0.184 & 0.084 & 1.10 & 8.08 
& 0.201 & 0.108 & 0.46 & 66.91 \\

$LDM_{CS_{25}}^{*}$ 
& 0.182 & 0.084 & 0.94 & 7.70
& 0.200 & 0.107 & 0.39 & 66.09 \\

$LDM_{CS_{30}}^{*}$ 
& 0.180 & 0.083 & 0.84 & 7.30 
& 0.199 & 0.107 & 0.35 & 65.63 \\

\bottomrule
\end{tabular}
\end{table}

\section{Conclusion}
\label{sec:conclusion}

We presented a streaming-aware framework for real-time video super-resolution that adapts pretrained single-image latent diffusion models through \emph{Cross-Step Attention} and \emph{Trajectory-Coupled Diffusion Scheduling}. By reusing intermediate denoising representations across adjacent frames and diffusion steps, the proposed approach enables implicit temporal information propagation without explicit motion estimation or dedicated temporal modules.

Experiments on REDS4 and YouHQ40-Test demonstrate improved perceptual quality and distribution-level temporal realism over vanilla latent diffusion, while lightweight fine-tuning further improves frame-wise temporal stability. \emph{Cross-Step Attention} also improves diffusion step efficiency, enabling strong perceptual quality with fewer denoising steps. Combined with the streaming inference pipeline, the proposed framework achieves over 40 FPS at $512^2$ after cold start, demonstrating the feasibility of real-time diffusion-based VSR.

Overall, our results show that exploiting the structure of the diffusion trajectory provides a simple and effective way to bridge high-quality diffusion restoration and real-time video processing.

\section*{Acknowledgements}
This work was funded by the European Union's Horizon Europe research and innovation programme under the Marie Sk\l{}odowska-Curie grant agreement No.~101129865 (ACCESS).

\bibliographystyle{plainnat}
\bibliography{references}




\end{document}